\documentclass[]{iflytek_robotics}

\usepackage{minitoc}
\usepackage{url}
\usepackage{amssymb}
\usepackage[utf8]{inputenc}
\usepackage{microtype}
\usepackage{booktabs}
\usepackage{pifont} 
\usepackage{multirow}
\usepackage{makecell}
\usepackage{paralist}
\usepackage{xspace}
\usepackage{color}
\usepackage{xcolor}
\usepackage{colortbl}
\usepackage{adjustbox}
\usepackage{hyperref} 
\usepackage[edges]{forest}
\usepackage{tikz} 
\usepackage{caption}
\usepackage{amsfonts}
\usepackage[toc,page,header]{appendix}
\usepackage[utf8]{inputenc}
\usepackage{bbm}
\usepackage{enumitem}

\definecolor{seedc}{RGB}{7, 92, 173}

\newcommand{\name}[1]{iFLYTEK-Embodied-Omni}

\renewcommand{\paragraph}[1]{\vspace{0.1em}\noindent\textbf{#1}}

\title{iFLYTEK-Embodied-Omni Technical Report}

\author[1]{Yuan Zhang} 
\author[1]{Jingfei Ni}
\author[3]{Guanchen Lu}
\author[3]{Shiqi Zhang}
\author[3]{Qingshan Xu}
\author[1]{Chi Liu}
\author[1]{Xin Nie}
\author[1]{Wenjie Xu}
\author[1]{Lin Gao}
\author[1]{Zhiyuan Cheng}
\author[1]{Mingxin Zhou}
\author[1]{Jiajia Wu}
\author[2]{Diyuan Liu}
\author[1,*]{Jia Pan}
\author[2,*]{Chao Ji}

\affiliation[1]{iFLYTEK}
\affiliation[2]{LindenBot}
\affiliation[3]{University of Science and Technology of China}
\abstract{
General-purpose embodied agents must understand multimodal instructions, anticipate how their environment will evolve, and produce precise control actions over extended horizons. Existing approaches typically specialize in visual-language reasoning, video-based world modeling, or action generation, while cascaded pipelines that first synthesize future observations and then infer actions can introduce interface bottlenecks and compound prediction errors. We present \textbf{iFLYTEK-Embodied-Omni}, a unified multimodal foundation model that jointly models vision (videos and images), language, and action within a single Omni framework. Its modality-specific visual-language, video-generation, and action-generation components communicate through shared multimodal self-attention. This design establishes brain--cerebellum collaboration: the vision-language model (VLM) and video generation model (VGM) form a high-level brain for instruction understanding, task planning, progress tracking, and future visual-state prediction, whereas the action generation model (AGM) serves as a low-level cerebellum that directly converts planned subgoals and shared multimodal context into executable action chunks. To develop these capabilities, we combine action-annotated and action-free embodied videos from human demonstrations and robot interactions with embodied reasoning, embodied perception, and general-purpose image-text data to construct a comprehensive dataset. We further adopt a four-stage strategy that progressively trains the VLM, VGM, and AGM before jointly fine-tuning the complete model. \textbf{iFLYTEK-Embodied-Omni} achieves an average success rate of 89.6\% on the zero-shot LIBERO-Plus benchmark, as well as 93.68\% and 93.16\% on the Clean and Rand settings of RoboTwin 2.0, respectively. Notably, on the RoboTwin 2.0 long-horizon subset, \textbf{iFLYTEK-Embodied-Omni} achieves the best average success rates of 88.3\% Clean and 89.0\% Rand, demonstrating the strong capabilities of the brain--cerebrum  collaborativ architecture in multi-stage planning, future-state anticipation, and temporally consistent action execution.
}

\date{\today}
\correspondence{\email{chaoji@iflytek.com, yuanzhang10@iflytek.com}}
\checkdata[Project Page]{\url{https://github.com/iFLYTEK-Embodied-Robotics-Team/iFLYTEK-Embodied-Omni}}

\begin{document}
\maketitle

\section{Introduction}
\label{sect:intro}

Building a general-purpose embodied foundation model requires more than mapping a visual observation and a language instruction directly to a short sequence of motor commands. A robot operating in an open-ended environment must understand the semantics of a task, recognize relevant objects and spatial relations \cite{han2022scene}, decompose a long-horizon objective into feasible stages \cite{ahn2022icanisay}, anticipate how the scene may change, and continually update its behavior according to execution progress \cite{huang2022innermonologueembodiedreasoning}. These capabilities are tightly coupled: task planning depends on perception, effective control depends on the intended future state, and progress tracking depends on comparing predicted outcomes with new observations. A broadly capable embodied intelligences should therefore reason across vision, language, environment dynamics, and action within a coherent computational framework rather than treating them as isolated problems.

Recent vision-language-action (VLA) models \cite{black2026pi0visionlanguageactionflowmodel, intelligence2025pi05visionlanguageactionmodelopenworld, kim2024openvlaopensourcevisionlanguageactionmodel, zheng2025xvlasoftpromptedtransformerscalable} leverage pretrained vision-language representations to equip robot policies with robust instruction understanding, semantic reasoning, and high-level task-planning capabilities. However, they typically learn action generation as a direct mapping from visual observations and language instructions to robot controls, without explicitly modeling how actions transform the environment. As a result, they have limited capacity to capture physical dynamics and anticipate future visual states. In contrast, world-action models (WAMs) \cite{li2026causalworldmodelingrobot, kim2026cosmospolicyfinetuningvideo, ye2026worldactionmodelszeroshot, zhu2025unifiedworldmodelscoupling} incorporate video-based world modeling to learn object motion, interaction outcomes, and spatiotemporal dynamics, providing a stronger foundation for predicting how the environment evolves. Nevertheless, existing WAMs predominantly focus on future visual prediction and low-level action generation, while offering comparatively limited support for language-grounded semantic reasoning, compositional task understanding, and long-horizon planning.

Unifying these complementary capabilities requires more than simply merging vision (images and videos), language, and action modalities. These modalities operate at different levels of semantic abstraction and temporal granularity and are optimized using heterogeneous learning objectives. Without explicit cross-modal alignment and modality-aware modeling, their representations may remain inconsistent, limiting knowledge transfer among semantic reasoning, world prediction, and action generation. Moreover, conventional WAMs \cite{du2023learninguniversalpoliciestextguided, hu2025videopredictionpolicygeneralist, pai2025mimicvideovideoactionmodelsgeneralizable} often connect world modeling and control through a cascaded pipeline that first generates future visual observations and then employs an inverse dynamics model to recover the corresponding actions. This explicit intermediate interface allows visual-generation errors to propagate to action prediction and prevents the action model from directly shaping representations for future prediction. Consequently, a central challenge is to jointly align heterogeneous modalities while combining the semantic intelligence of VLAs with the dynamics-modeling capabilities of WAMs.

To address these challenges, we present \textbf{iFLYTEK-Embodied-Omni}, a unified multimodal foundation model that jointly models vision (images and videos), language, and action within a single Omni framework. Rather than naively merging heterogeneous modalities, our model preserves modality-specific representations while enabling cross-modal alignment and information exchange through shared multimodal self-attention. The vision language model (VLM) and video generation model (VGM) form a high-level ``brain'' for semantic reasoning, task planning, progress tracking, and future visual prediction, whereas the action generation model (AGM) serves as a low-level ``cerebellum'' that translates planned subgoals into executable action chunks. Unlike cascaded video-generation-to-action pipelines, the AGM directly generates actions from the shared multimodal context, tightly coupling semantic intent, predicted world evolution, and robot control.

We train \textbf{iFLYTEK-Embodied-Omni} using a multi-source mixture of action-annotated and action-free embodied videos, embodied reasoning and perception data, and general-purpose image-text data. Our four-stage strategy progressively specializes the VLM, VGM, and AGM before jointly fine-tuning the complete model, thereby aligning multimodal understanding, world modeling, and action generation. On the zero-shot LIBERO-Plus benchmark, iFlytek-Embodied-Omni achieves an average success rate of 89.6\%; it further reaches 93.68\% and 93.16\% under the Clean and Rand settings of RoboTwin 2.0, respectively, which demonstrates the strong capabilities of the brain--cerebrum collaborative architecture in multi-stage planning, future-state anticipation, and temporally consistent action execution.

Our contributions are summarized as follows:
\begin{itemize}
    \item We present \textbf{iFLYTEK-Embodied-Omni}, a unified multimodal foundation model that jointly models vision (images and videos), language, and action within a single Omni framework. It supports multimodal understanding and generation, future visual-state prediction, and embodied action generation.

    \item We establish a brain--cerebellum collaborative architecture within the unified model. The VLM and VGM serve as the brain to understand tasks, formulate plans, predict future visual states, and track task progress, while the AGM acts as the cerebellum to translate planned subgoals into executable action chunks for coordinated long-horizon task execution.

    \item We construct a multi-source training mixture comprising action-annotated and action-free embodied video data from human demonstrations and robot interactions, embodied reasoning data, embodied perception data, and general-purpose image-text data. This mixture connects general multimodal knowledge with embodied perception, reasoning, world modeling, and action learning.

    \item We develop a four-stage training strategy comprising VLM fine-tuning, VGM training through future video prediction, AGM training with the VLM and VGM frozen, and joint fine-tuning of all three components. This progressive-to-joint optimization aligns multimodal understanding, visual world modeling, and action generation for long-horizon embodied control.
\end{itemize}

\section{Overview}
\label{sect:overview}

\begin{figure}[ht]
    \centering
    \includegraphics[width=\textwidth]{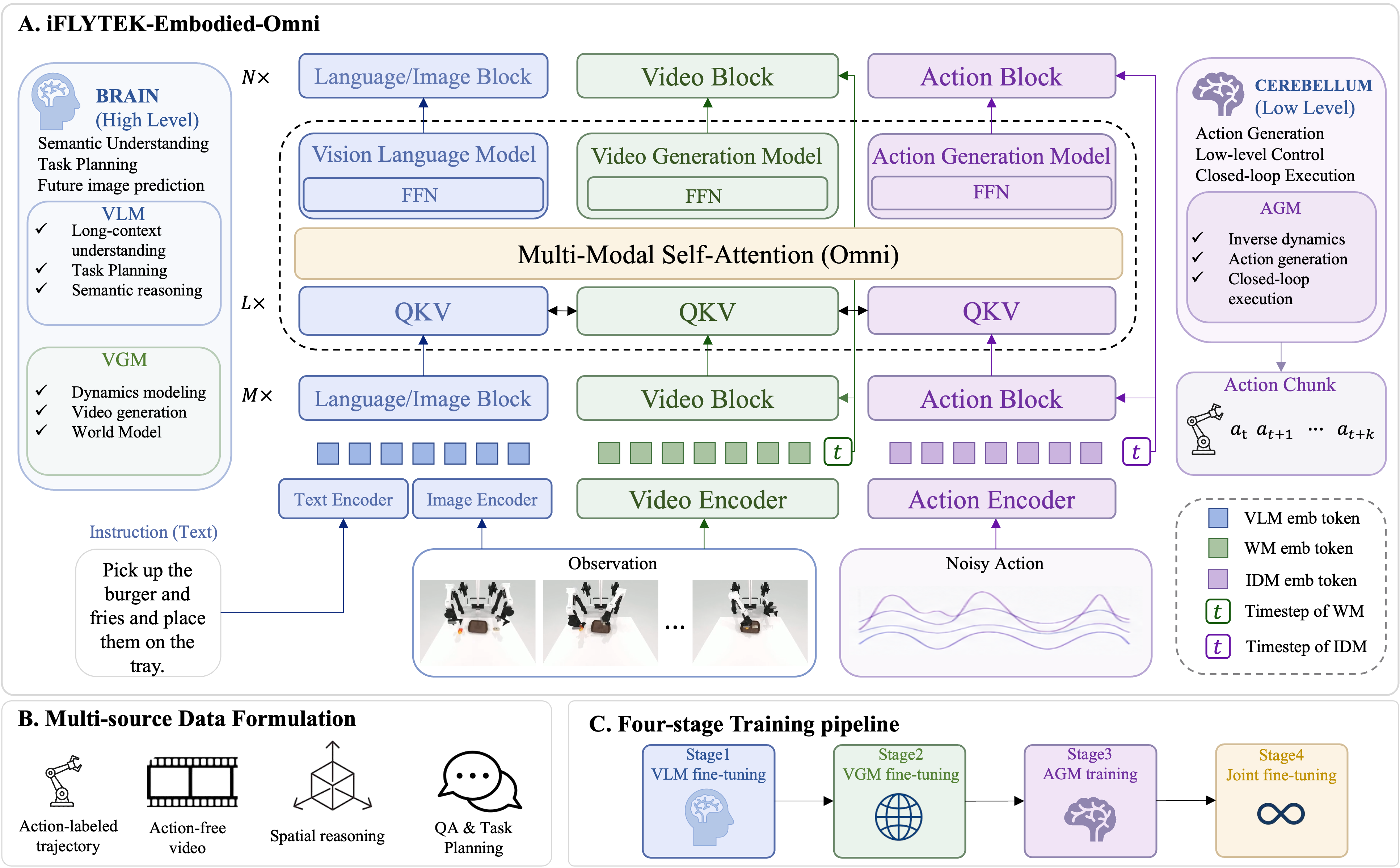} % 插入图片，并设置宽度为文本宽度的50%
    \caption{
    \textbf{Overview of iFLYTEK-Embodied-Omni.}
    (A) The unified Omni architecture aligns language, vision(image and video), and action representations through modality-specific blocks and shared multimodal self-attention. The VLM and VGM form the high-level ``brain,'' while the AGM serves as the low-level ``cerebellum'' for action generation.
    (B) Multi-source training data include action-labeled trajectories, action-free videos, spatial-reasoning data, and question-answering and task-planning data.
    (C) The four-stage pipeline progressively trains the VLM, VGM, and AGM, followed by joint fine-tuning of the complete model.
    } % 图片标题
    \label{fig:framework} % 创建标签，用于引用
\end{figure}

Figure~\ref{fig:framework} presents an overview of iFLYTEK-Embodied-Omni and its training pipeline. The framework contains three closely connected parts: (A) a unified brain--cerebellum Omni architecture that aligns vision (images and videos), language, and action, coordinating high-level multimodal reasoning and world modeling with low-level action generation; (B) a multi-source data formulation that combines general multimodal knowledge with embodied experience; and (C) a four-stage pipeline that progressively develops and integrates the model's visual-language, world-modeling, and action-generation capabilities.

\paragraph{A. Unified Omni architecture.}
iFLYTEK-Embodied-Omni is a unified multimodal foundation model that jointly models vision (images and videos), language, and action within a single Omni framework. By aligning these modalities in a shared context, the model supports multimodal understanding and generation, future visual-state prediction, and embodied action generation. Within this unified framework, the VLM and VGM form a high-level ``brain'' that understands tasks, formulates plans, predicts future visual states, and tracks task progress. The AGM serves as a low-level ``cerebellum'' that translates planned subgoals into executable action chunks. This brain--cerebellum collaboration coordinates high-level reasoning, world modeling, and low-level control for long-horizon task execution.

\paragraph{B. Multi-source data formulation.}
As illustrated in Fig.~\ref{fig:framework}(B), the training mixture comprises four complementary sources. Action-labeled robot trajectories provide direct supervision for executable control, while action-free videos from human demonstrations and robot interactions supply broader priors about motion and physical dynamics. Embodied perception and spatial-reasoning data strengthen object-centric and geometric understanding. General image-text, question-answering, and task-planning data further develop semantic knowledge, instruction comprehension, and compositional reasoning. This mixture connects general multimodal and video priors with embodied reasoning and robot action spaces.

\paragraph{C. Four-stage training pipeline.}
We use the staged-to-joint pipeline shown in Fig.~\ref{fig:framework}(C) to reduce interference among heterogeneous learning objectives while ultimately aligning all model components.

1) Stage I: We first fine-tune the VLM to establish embodied visual-language understanding and generation, embodied semantic reasoning, embodied spatial perception, and embodied task-planning capabilities.

2) Stage II: We then fine-tune the VGM on embodied video prediction, enabling it to model embodied spatiotemporal dynamics and anticipate future visual states within the shared Omni context.

3) Stage III: With the VLM and VGM frozen, we train the AGM on action-labeled robot trajectories. This stage aligns the representations produced by the high-level brain with executable action chunks without disrupting previously acquired semantic and dynamics knowledge.

4) Stage IV: Finally, we jointly fine-tune the VLM, VGM, and AGM, allowing semantic reasoning, future-state prediction, and action generation to mutually adapt within the unified model. This progressive-to-joint strategy first develops specialized capabilities and subsequently coordinates them for closed-loop embodied control.

The following sections describe the Omni architecture and its modality-specific components in detail, followed by the data construction and optimization objectives used in each training stage.

\section{\name{} Model}
\label{sect:model}

iFLYTEK-Embodied-Omni is a unified multimodal foundation model that models vision (videos and images), language, and action within a single Omni framework. Given a language instruction, visual observations, video context, and noisy action tokens, the model produces multimodal outputs including visual-language responses, future visual states, and executable action chunks. Instead of treating semantic understanding, world modeling, and action generation as isolated modules, iFLYTEK-Embodied-Omni aligns them through a shared multimodal context. This section introduces the model from three aspects: the Omni architecture and brain--cerebellum mechanism, the cross-modal attention mask and image positional encoding, and the inference procedure for closed-loop embodied control.

\subsection{Model Architecture}

The core architecture of iFLYTEK-Embodied-Omni consists of three modality-specialized components: a vision-language model (VLM), a video generation model (VGM), and an action generation model (AGM). The VLM is inherited from a pretrained vision-language model and provides language-image understanding, semantic reasoning, and task planning. The VGM is inherited from a pretrained video diffusion model and learns visual world dynamics through future visual-state prediction. The AGM is initialized from the video diffusion weights of the VGM and further fine-tuned on robot trajectory data, allowing the action model to reuse temporal generation priors while adapting them to executable robot control.

As shown in Fig.~\ref{fig:framework}(A), each modality stream is first processed by its own input encoder and modality-specific Transformer blocks. The language-image, video, and action features are independently passed through $M$ blocks to perform inner self-attention within each modality. The resulting tokens are then concatenated and fed into $L$ Omni Multi-Modal Self-Attention layers, where cross-modal self-attention aligns vision (images and videos), language, and action representations in a shared context. After this Omni interaction stage, the token streams are separated again and processed by $N$ modality-specific blocks, enabling the VLM, VGM, and AGM to perform their respective output modeling while benefiting from the aligned multimodal context.

This structure naturally forms a brain--cerebellum collaboration. The VLM and VGM serve as the high-level ``brain'', which decomposes tasks, performs spatial and semantic reasoning, predicts future visual states, and tracks task progress. The AGM acts as the low-level ``cerebellum'', which uses the subgoal information produced by the brain and the shared Omni context to generate executable action chunks.

During closed-loop execution, the newly observed image after action execution replaces the previously predicted visual state in a teacher-forcing manner, allowing the model to continually correct its internal context with real environment feedback. Through Omni-based cross-modal alignment, iFLYTEK-Embodied-Omni supports vision-language modeling, vision-language-action policy modeling, world modeling, video generation, inverse dynamics, and joint video-action prediction within a unified model, which are formulated in Table~\ref{tab:prediction_modes}.

\begin{table}[t]
    \centering
    \caption{Different prediction modes of iFLYTEK-Embodied-Omni.}
    \label{tab:prediction_modes}
    {\footnotesize
    \setlength{\tabcolsep}{4pt}
    \renewcommand{\arraystretch}{1.05}
    \begin{tabular}{c|c}
        \hline
        Model & Prediction Objective \\
        \hline
        VLM & $p(\ell_{goal} \mid \mathbf{o}_t, \ell)$ \\
        VLA & $p(\mathbf{a}_{t+1:t+k} \mid \mathbf{o}_t, \ell)$ \\
        WM & $p(\mathbf{o}_{t+1:t+k} \mid \mathbf{o}_t, \mathbf{a}_{t+1:t+k})$ \\
        IDM & $p(\mathbf{a}_{t+1:t+k} \mid \mathbf{o}_{t:t+k})$ \\
        VGM & $p(\mathbf{o}_{t+1:t+k} \mid \mathbf{o}_t, \ell)$ \\
        Joint Video-Action Prediction & $p(\mathbf{o}_{t+1:t+k}, \mathbf{a}_{t+1:t+k} \mid \mathbf{o}_t, \ell)$ \\
        \hline
    \end{tabular}}
\end{table}

\subsection{Cross-Modal Attention Mask and Image Positional Encoding}

In the Omni Multi-Modal Self-Attention layers, iFLYTEK-Embodied-Omni jointly processes four types of modality tokens: image, language, video, and action. Since these modalities follow different generation orders and prediction objectives, we do not apply a single uniform attention pattern to all tokens. Instead, each prediction mode in Table~\ref{tab:prediction_modes} is associated with a modality-aware attention mask that specifies which tokens can be attended to during training and inference.

\begin{figure*}[t]
    \centering
    \includegraphics[width=\textwidth]{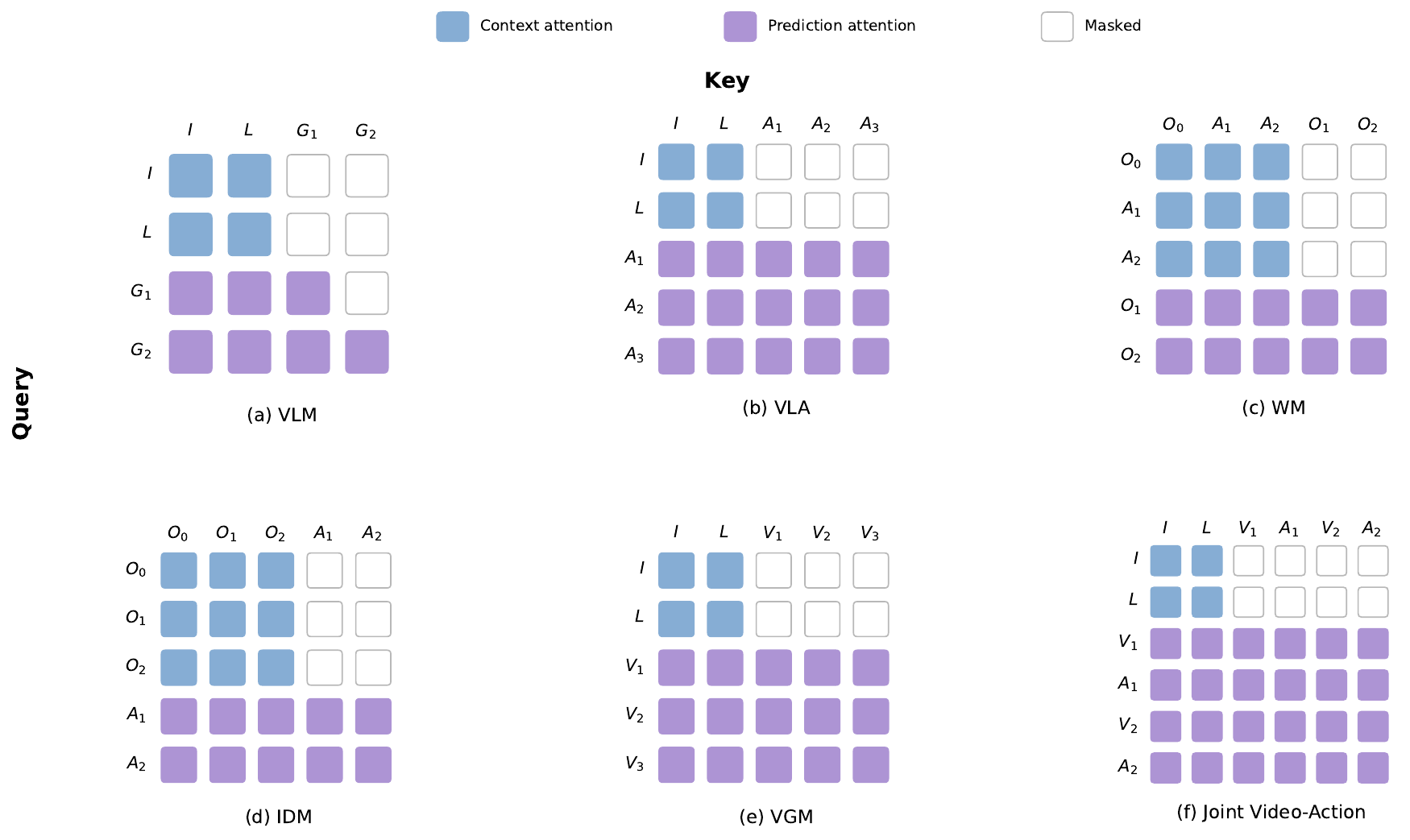}
    \caption{
    Attention masks for different prediction modes in iFLYTEK-Embodied-Omni.
    Colored entries denote visible attention connections, while empty entries denote masked connections.
    Blue entries correspond to conditioning/context attention, and purple entries correspond to prediction-side attention.
    }
    \label{fig:attention_masks}
\end{figure*}

For image-language modeling, image and language tokens are always allowed to interact with each other. Image tokens within the same image use bidirectional attention to preserve global visual context, while language tokens follow an autoregressive causal mask within the text sequence. For video modeling, video tokens use bidirectional attention both within each frame and across frames, enabling the VGM to model spatial structure and temporal dynamics jointly. For action modeling, all action tokens within an action chunk use bidirectional attention, allowing the AGM to generate temporally coherent low-level controls. Across different prediction modes, the mask exposes only the valid conditioning context required by the corresponding objective while preventing target tokens from leaking unavailable future information, as shown in Fig.~\ref{fig:attention_masks}.

For multi-view video inputs, each timestamp may contain observations from multiple camera views. We first encode each view with the VAE encoder and concatenate the resulting image latents as the video-token representation at that timestamp. Within each image latent, we use RoPE to encode spatial positions. To distinguish different camera views, we assign each view an independent learnable view embedding and add it to the corresponding latent tokens. This combination preserves intra-view spatial structure while allowing the model to identify cross-view correspondences during video prediction, world modeling, and action generation.

\subsection{Inference Process}

At inference time, iFLYTEK-Embodied-Omni receives the current language instruction, visual observation, and robot state, and generates an executable action chunk through iterative denoising. The VLM encodes the instruction and observation into semantic planning context, while the VGM provides visual-dynamics context for future-state reasoning. The AGM then refines noisy action latents into low-level controls conditioned on the shared Omni context. After each action chunk is executed, the newly captured observation is fed back to the model, replacing the previously predicted visual state and enabling closed-loop replanning.

To reduce the cost of iterative denoising, we introduce a DiT velocity cache that exploits the temporal smoothness of consecutive denoising predictions. Let $\mathbf{u}^{(s)}$ denote the velocity predicted by the DiT module at denoising step $s$, where $\mathbf{u}$ can correspond to either video velocity or action velocity depending on the inference mode. We compute the cosine consistency between two adjacent predictions as
\begin{equation}
    \rho_s =
    \frac{
        \left\langle \mathbf{u}^{(s)}, \mathbf{u}^{(s-1)} \right\rangle
    }{
        \left\|\mathbf{u}^{(s)}\right\|_2
        \left\|\mathbf{u}^{(s-1)}\right\|_2
    } .
\end{equation}
When $\rho_s$ exceeds a threshold $\gamma$, the following $c$ denoising evaluations are skipped and their velocities are approximated using the most recent reliable prediction:
\begin{equation}
    \widehat{\mathbf{u}}^{(s+r)} = \mathbf{u}^{(s)}, \quad r = 1,\ldots,c .
\end{equation}
The cache is cleared at the beginning of each inference call or action chunk, so that the approximation only reuses local denoising history within the current sampling trajectory.

We further adopt a V2A-style inference schedule for models trained with asymmetric video-to-action attention. Under this attention pattern, video tokens do not attend to action tokens, whereas action tokens can attend to language and visual context. This allows us to avoid repeatedly updating the video branch throughout the whole sampling process. Let $\mathbf{z}^{(s)}_v$ and $\mathbf{z}^{(s)}_a$ denote the video and action latents at denoising step $s$, respectively. We first run a short joint denoising prefix for $N$ steps, and then freeze the video latent while continuing action-only denoising:
\begin{equation}
\left(\mathbf{z}^{(s+1)}_v, \mathbf{z}^{(s+1)}_a\right)=
\begin{cases}
    \Phi_{\mathrm{joint}}\!\left(\mathbf{z}^{(s)}_v,\mathbf{z}^{(s)}_a,s\right), & s < N, \\
    \left(\mathbf{z}^{(N)}_v,
    \Phi_{\mathrm{act}}\!\left(\mathbf{z}^{(s)}_a;\mathbf{z}^{(N)}_v,s\right)\right), & s \geq N .
\end{cases}
\end{equation}
After the prefix stage, the fixed video-language branch is executed once to construct cached key-value states for the visual and textual context. Subsequent denoising steps update only action tokens: action queries attend to the cached video-language keys and values together with the current action keys and values. This schedule preserves the attention semantics of V2A-style modeling while removing redundant video-stream computation from the later sampling steps.

Combining DiT velocity caching with V2A-style staged denoising yields a lightweight closed-loop inference procedure. The model repeatedly observes the environment, constructs high-level semantic and visual-dynamics context, generates an action chunk, executes it, and refreshes the context with the latest observation. In our implementation, these optimizations enable iFLYTEK-Embodied-Omni to achieve an inference speed of 3 Hz on a single RTX 4090 GPU while maintaining a 30 Hz robot control frequency.

\section{Training Strategy}
\label{sect:strategy}

\subsection{Data Preparation}
\label{sec:data_preparation}

\begin{figure*}[t]
    \centering
    \includegraphics[width=\textwidth]{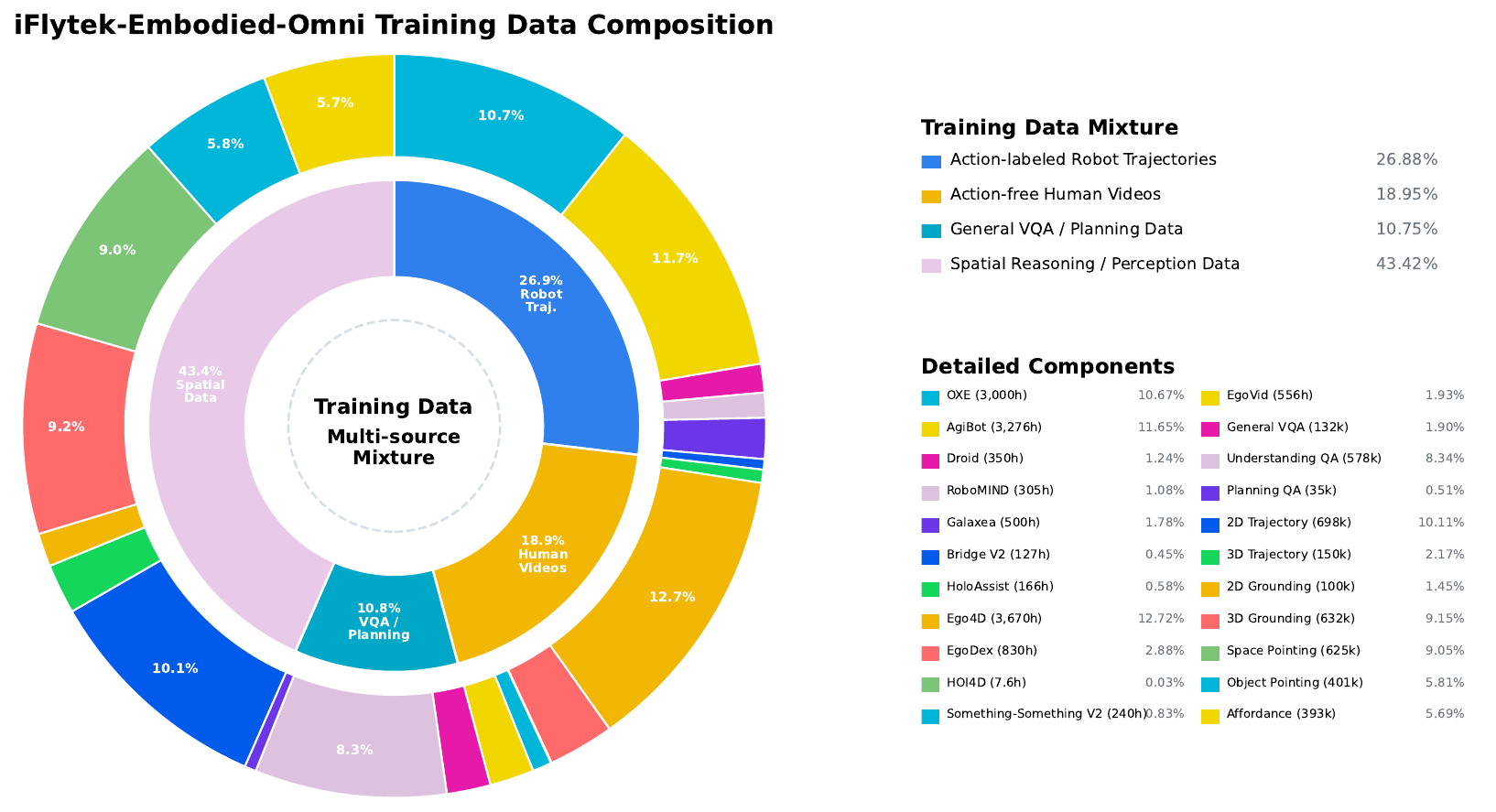}
    \caption{
    Training data composition of iFLYTEK-Embodied-Omni.
    The inner ring shows the four major data categories, while the outer ring details the datasets used in each category.
    }
    \label{fig:dataset_composition}
\end{figure*}

Training iFLYTEK-Embodied-Omni requires data that cover both general multimodal intelligence and embodied interaction. Since the model jointly learns visual-language understanding, visual world dynamics, and executable action generation, relying only on robot trajectories would provide limited diversity in semantic reasoning and physical interactions. We therefore construct a multi-source training mixture consisting of four complementary data categories, as illustrated in Fig.~\ref{fig:dataset_composition}.

First, we use open-source action-labeled robot trajectory datasets, which include OXE \cite{embodimentcollaboration2025openxembodimentroboticlearning}, AgiBot \cite{agibotworldcontributors2025agibotworldcolosseolargescale}, Droid \cite{khazatsky2025droidlargescaleinthewildrobot}, RoboMIND \cite{Wu_2025}, Galaxea \cite{jiang2025galaxeaopenworlddatasetg0}, and Bridge V2 \cite{walke2024bridgedatav2datasetrobot}, to provide direct supervision for robot control and action generation. These datasets cover diverse robot embodiments, manipulation scenes, and single-arm or dual-arm control settings, enabling the AGM to learn executable action patterns from large-scale robot experience.

Second, we incorporate open-source human operation videos which include action-free datasets to improve the model's ability to understand manipulation dynamics beyond robot-only demonstrations. This category includes HoloAssist \cite{wang2023holoassistegocentrichumaninteraction}, Ego4D \cite{grauman2022ego4dworld3000hours}, EgoDex \cite{hoque2026egodexlearningdexterousmanipulation}, HOI4D \cite{liu2024hoi4d4degocentricdataset}, Something-Something V2 \cite{goyal2017somethingsomethingvideodatabase}, and EgoVid \cite{wang2024egovid5mlargescalevideoactiondataset}. Although these videos do not provide robot action labels, they contain rich object interactions, hand-object motion, and temporal changes, which are useful for training the VGM to acquire visual dynamics and future-state prediction capability.

Third, we build in-house general image-language understanding and task-planning datasets, including General VQA with 132K samples, Understanding QA with 578K samples, and Planning QA with 35K samples. These data are used to strengthen visual-language understanding, instruction following, task decomposition, and semantic planning, providing the VLM with the high-level reasoning capability required by long-horizon embodied tasks.

Finally, we construct in-house spatial reasoning and embodied perception datasets to enhance object-centric and geometry-aware understanding. This category contains 2D Trajectory with 698K samples, 3D Trajectory with 150K samples, 2D Grounding with 100K samples, 3D Grounding with 632K samples, Space Pointing with 625K samples, Object Pointing with 401K samples, and Affordance with 393K samples. These data improve spatial localization, trajectory reasoning, grounding, pointing, and affordance perception, which are essential for aligning language instructions with actionable visual scenes.

As shown in Fig.~\ref{fig:dataset_composition}, action-labeled robot trajectories account for 26.88\% of the full training mixture, action-free human operation videos account for 18.95\%, embodied general VQA and task-planning data account for 10.75\%, and embodied spatial reasoning and embodied perception data account for 43.42\%. This data composition allows iFLYTEK-Embodied-Omni to connect general multimodal knowledge, visual dynamics, spatial reasoning, and robot action spaces within a unified training framework.

\subsection{Training Process}
\label{sec:training_process}

We train iFLYTEK-Embodied-Omni with a four-stage progressive-to-joint strategy. The first three stages develop visual-language understanding, visual world modeling, and action generation in sequence, reducing interference among heterogeneous objectives. The final stage jointly optimizes the complete model to align these capabilities within the shared Omni context.

\noindent 1) Stage I: VLM Fine-tuning.
We first fine-tune the VLM using the general image-text, embodied perception, spatial-reasoning, and task-planning data described in Sec.~\ref{sec:data_preparation}. Given an observation $\mathbf{o}_t$, an instruction $\ell$, and a target response $\mathbf{y}=(y_1,\ldots,y_T)$, the VLM is optimized with an autoregressive language-modeling objective:
\begin{equation}
    \mathcal{L}_{\mathrm{VLM}}
    =
    -\mathbb{E}_{(\mathbf{o}_t,\ell,\mathbf{y})}
    \left[
        \sum_{i=1}^{T}
        \log p_{\theta_V}
        \left(y_i \mid y_{<i},\mathbf{o}_t,\ell\right)
    \right],
\end{equation}
where $\theta_V$ denotes the VLM parameters. This stage establishes instruction understanding, embodied perception, spatial reasoning, task decomposition, and visual-language generation before introducing the video and action objectives.

\noindent 2) Stage II: VGM Training.
We next freeze the VLM and train the VGM to predict future visual states from the current observation and language context. Let $\mathbf{z}_v$ denote the VAE latent representation of a target future video and let $\boldsymbol{\epsilon}_v\sim\mathcal{N}(\mathbf{0},\mathbf{I})$ be Gaussian noise. At noise level $\sigma_v\in[0,1]$, the flow-matching path is defined as
\begin{equation}
    \mathbf{z}^{\sigma_v}_v
    =
    (1-\sigma_v)\mathbf{z}_v
    +
    \sigma_v\boldsymbol{\epsilon}_v,
    \qquad
    \mathbf{u}^{*}_v
    =
    \boldsymbol{\epsilon}_v-\mathbf{z}_v .
\end{equation}
The VGM learns the corresponding velocity field through
\begin{equation}
    \mathcal{L}_{\mathrm{VGM}}
    =
    \mathbb{E}
    \left[
        \left\|
        \mathbf{v}^{v}_{\theta_G}
        \left(
            \mathbf{z}^{\sigma_v}_v,
            \mathbf{o}_t,
            \ell,
            \sigma_v
        \right)
        -
        \mathbf{u}^{*}_v
        \right\|_2^2
    \right],
\end{equation}
where $\theta_G$ denotes the VGM parameters. Future-video supervision allows the model to acquire object-motion, interaction, and scene-evolution priors while retaining the semantic context supplied by the frozen VLM.

\noindent 3) Stage III: AGM Training.
After world-model training, we freeze both the VLM and VGM and optimize the AGM on action-annotated robot trajectories. The AGM is initialized from the VGM weights so that temporal generation priors can be adapted to continuous robot control. For an action chunk represented by $\mathbf{z}_a$, we independently sample $\boldsymbol{\epsilon}_a\sim\mathcal{N}(\mathbf{0},\mathbf{I})$ and construct
\begin{equation}
    \mathbf{z}^{\sigma_a}_a
    =
    (1-\sigma_a)\mathbf{z}_a
    +
    \sigma_a\boldsymbol{\epsilon}_a,
    \qquad
    \mathbf{u}^{*}_a
    =
    \boldsymbol{\epsilon}_a-\mathbf{z}_a .
\end{equation}
Conditioned on the frozen brain context, the AGM predicts the action velocity field. Its training objective is
\begin{equation}
    \mathcal{L}_{\mathrm{AGM}}
    =
    \mathbb{E}
    \left[
        \frac{
            \left\|
            \mathbf{m}_a\odot
            \left(
                \mathbf{v}^{a}_{\theta_A}
                \left(
                    \mathbf{z}^{\sigma_a}_a,
                    \mathbf{o}_t,
                    \ell,
                    \sigma_a
                \right)
                -
                \mathbf{u}^{*}_a
            \right)
            \right\|_2^2
        }{
            \|\mathbf{m}_a\|_1+\varepsilon
        }
    \right],
\end{equation}
where $\theta_A$ denotes the AGM parameters, $\mathbf{m}_a$ masks padded or invalid action dimensions, and $\varepsilon$ prevents numerical instability. Freezing the brain components at this stage prevents action-learning gradients from disrupting their previously acquired semantic and dynamics representations.

\paragraph{Modality-specific noise sampling.}
Video and action generation have different sensitivity to corruption, so we use separate SNR-shifted timestep distributions for their flow-matching objectives. For modality $m\in\{v,a\}$, we first draw $r_m\sim\mathcal{U}(0,1)$ and transform it into a noise level by
\begin{equation}
    \sigma_m
    =
    \frac{s_m r_m}
    {1+(s_m-1)r_m}.
\end{equation}
We set the shift factors to $s_v=6$ for video and $s_a=1$ for action. The larger video shift places more training mass in highly corrupted regions, whereas the action noise levels remain uniformly distributed. Consequently, the AGM learns to recover accurate controls even when the visual dynamics context is uncertain or noisy. This asymmetric sampling improves optimization stability and permits the number of sampling steps at inference to be reduced from 50 to 30 without degrading control performance.

\noindent 4) Stage IV: Joint Fine-tuning.
Finally, we unfreeze the VLM, VGM, AGM, and shared Omni layers and jointly fine-tune the complete model. Each sample activates the objectives associated with its available supervision, and the overall loss is written as
\begin{equation}
    \mathcal{L}_{\mathrm{joint}}
    =
    \lambda_{\mathrm{vlm}}\mathcal{L}_{\mathrm{VLM}}
    +
    \lambda_{\mathrm{vgm}}\mathcal{L}_{\mathrm{VGM}}
    +
    \lambda_{\mathrm{agm}}\mathcal{L}_{\mathrm{AGM}},
\end{equation}
where $\lambda_{\mathrm{vlm}}$, $\lambda_{\mathrm{vgm}}$, and $\lambda_{\mathrm{agm}}$ balance the three training objectives. This final stage allows semantic reasoning, future visual-state prediction, and action generation to mutually adapt through Omni Multi-Modal Self-Attention, completing the alignment between the high-level brain and low-level cerebellum for long-horizon embodied control.

\section{Experiments}
\label{sect:experiments}

To systematically evaluate iFLYTEK-Embodied-Omni, we conduct four complementary experiments in simulated environments. We first evaluate zero-shot generalization on LIBERO-Plus across variations in camera viewpoint, robot embodiment, language instruction, illumination, background, visual noise, and scene layout. We then evaluate the model on the Clean and Rand settings of RoboTwin 2.0 to examine its manipulation performance and robustness under randomized environments. Together, these two benchmarks provide a comprehensive assessment of general-purpose policy learning and cross-domain generalization.

Beyond aggregate benchmark performance, we conduct targeted comparisons on long-horizon tasks that require multi-stage planning, future-state reasoning, task-progress tracking, and closed-loop execution. Finally, we perform ablation studies to investigate the effectiveness of the unified three-branch architecture and analyze the respective roles of the VLM, VGM, and AGM. These experiments examine whether semantic reasoning, visual world modeling, and action generation provide complementary capabilities when coordinated within the unified Omni framework.

\subsection{Simulation Experiments on LIBERO-Plus}
\label{sec:libero_plus}

We evaluate the zero-shot generalization capability of iFLYTEK-Embodied-Omni on LIBERO-Plus. The benchmark introduces seven types of distribution shifts into robot manipulation tasks, covering changes in camera viewpoint, robot embodiment, language instruction, illumination, background, visual noise, and scene layout. We report the success rate under each variation and the average success rate over the complete evaluation set. This protocol examines whether a policy can preserve task understanding and executable control when the visual, linguistic, and embodiment conditions differ from those encountered during training.

\begin{table*}[t]
    \centering
    \caption{
    Zero-shot success rates (\%) on LIBERO-Plus.
    \textbf{Bold} and \underline{underlined} values indicate the best and second-best results in each column, respectively.
    Ave. denotes the average success rate computed over the complete evaluation set.
    }
    \label{tab:libero_plus}
    \scriptsize
    \setlength{\tabcolsep}{4pt}
    \renewcommand{\arraystretch}{1.08}
    \resizebox{\textwidth}{!}{
    \begin{tabular}{lcccccccc}
        \hline
        Method &
        Camera &
        Robot &
        Language &
        Light &
        Background &
        Noise &
        Layout &
        Ave. \\
        \hline
        \multicolumn{9}{c}{\textit{VLA-based methods}} \\
        \hline
        OpenVLA       & 0.8  & 3.5  & 23.0 & 8.1  & 34.8 & 15.2 & 28.5 & 15.6 \\
        OpenVLA-OFT   & 56.4 & 31.9 & 79.5 & 88.7 & 93.3 & 75.8 & 74.2 & 69.6 \\
        UniVLA        & 1.8  & 46.2 & 69.6 & 69.0 & 81.0 & 21.2 & 31.9 & 42.9 \\
        NORA          & 2.2  & 37.0 & 65.1 & 45.7 & 58.6 & 12.8 & 62.1 & 39.0 \\
        $\pi_0$-Fast  & 65.1 & 21.6 & 61.0 & 73.2 & 73.2 & 74.4 & 68.8 & 61.6 \\
        $\pi_0$       & 61.0 & 40.8 & 63.5 & 89.3 & 84.1 & 80.1 & 76.4 & 69.4 \\
        $\pi_{0.5}$   & 75.8 & 79.4 & 83.3 & 95.5 & 95.0 & 89.6 & 87.0 & 85.7 \\
        ACoT          & 72.6 & 82.6 & \textbf{87.5} & 97.7 & \textbf{96.5} & 87.8 & 88.1 & \underline{86.6} \\
        \hline
        \multicolumn{9}{c}{\textit{WAM-based methods}} \\
        \hline
        WorldVLA      & 0.1  & 27.9 & 41.6 & 43.7 & 17.1 & 10.9 & 38.0 & 25.0 \\
        Cosmos-Policy & 75.8 & 63.3 & 81.7 & 96.5 & 88.9 & 92.7 & 82.2 & 82.2 \\
        HoloBrain-0   & 65.5 & 58.2 & 78.7 & 88.1 & 90.3 & 66.9 & 79.5 & 74.0 \\
        OA-WAM        & 80.5 & \textbf{89.6} & 85.3 & 96.5 & 95.9 & 75.6 & 82.8 & 83.9 \\
        World Pilot   & \underline{82.8} & 60.6 & \underline{87.2} & \textbf{98.6} & \underline{96.4} & \underline{93.6} & 80.5 & 84.7 \\
        CKT-WAM       & 77.4 & 71.4 & 86.7 & 98.2 & 90.2 & \textbf{94.8} & \underline{88.5} & 86.1 \\
        \textbf{iFLYTEK-Embodied-Omni} &
        \textbf{84.8} &
        \underline{85.5} &
        84.6 &
        \underline{98.4} &
        94.3 &
        92.4 &
        \textbf{91.2} &
        \textbf{89.6} \\
        \hline
    \end{tabular}}
\end{table*}

As reported in Table~\ref{tab:libero_plus}, iFLYTEK-Embodied-Omni achieves the highest average success rate of 89.6\%. It exceeds the strongest overall baseline, ACoT, by 3.0 percentage points and the strongest competing WAM method, CKT-WAM, by 3.5 percentage points. The model obtains the best performance under camera and layout variations, reaching 84.8\% and 91.2\%, respectively. These results improve over the corresponding second-best methods by 2.0 and 2.7 percentage points, indicating strong robustness to viewpoint changes and spatial rearrangements.

The model also ranks second under robot and lighting variations, with success rates of 85.5\% and 98.4\%. Under language, background, and noise shifts, it achieves 84.6\%, 94.3\%, and 92.4\%, respectively. Although these individual results are not the highest in every category, the consistently strong performance across all seven variations produces the best overall average. This balanced behavior is consistent with the design goal of coordinating semantic reasoning, visual world modeling, and action generation within a unified Omni context, rather than optimizing for only one type of environmental variation.

\subsection{Simulation Experiments on RoboTwin 2.0}
\label{sec:robotwin2}

We further evaluate iFLYTEK-Embodied-Omni on RoboTwin 2.0 using its Clean and Rand settings. Clean Avg measures average task success under the standard simulation configuration, whereas Rand Avg evaluates generalization and robustness when the environment is randomized. We report the average success rate for both settings and compare against representative VLA-based and WAM-based approaches.

\begin{figure*}[t]
    \centering
    \includegraphics[width=\textwidth]{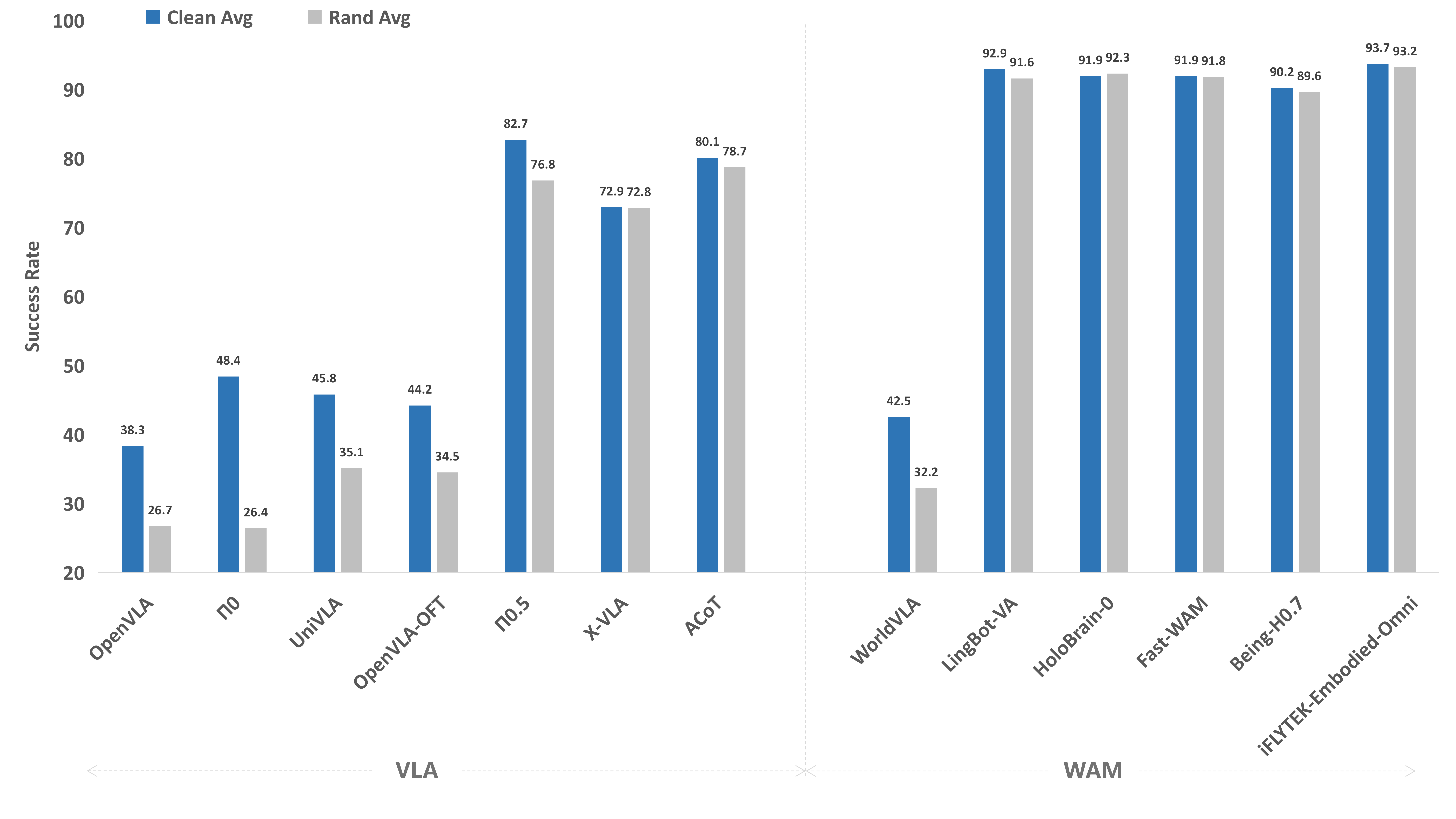}
    \caption{
    Average success rates on the Clean and Rand settings of RoboTwin 2.0.
    Methods are grouped into VLA-based and WAM-based approaches.
    Bar labels are rounded to one decimal place for visualization.
    }
    \label{fig:robotwin2_results}
\end{figure*}

As shown in Fig.~\ref{fig:robotwin2_results}, iFLYTEK-Embodied-Omni achieves the highest average success rate under both evaluation settings, reaching 93.68\% on Clean and 93.16\% on Rand. On the Clean setting, it exceeds the second-best method, LingBot-VA at 92.93\%, by 0.75 percentage points. Under the more challenging Rand setting, it outperforms HoloBrain-0 at 92.30\% by 0.86 percentage points.

The performance decreases by only 0.52 percentage points from Clean to Rand, indicating that the policy remains stable under environment randomization. Compared with the strongest VLA baselines, iFLYTEK-Embodied-Omni improves over $\pi_{0.5}$ by 10.94 percentage points on Clean and over ACoT by 14.44 percentage points on Rand. Together with its competitive performance against recent WAM approaches, these results are consistent with the benefit of coordinating semantic reasoning, visual world modeling, and action generation within a shared Omni context.

\subsection{Long-Horizon Task Evaluation}
\label{sec:long_horizon}

To further examine long-horizon embodied control, we select seven multi-stage tasks from RoboTwin 2.0: Put Bottles in Dustbin, Open Microwave, Rank Blocks by Size, Stack Three Blocks, Stack Three Bowls, Hang Mug, and Handover Block. These tasks cover object sorting, sequential stacking, container manipulation, articulated-object interaction, and object handover. Successful execution requires the policy to preserve task intent across multiple action chunks, track intermediate progress, and recover from state changes during execution. We report the success rate under both the Clean and Rand settings.

\begin{figure*}[t]
    \centering
    \includegraphics[width=\textwidth]{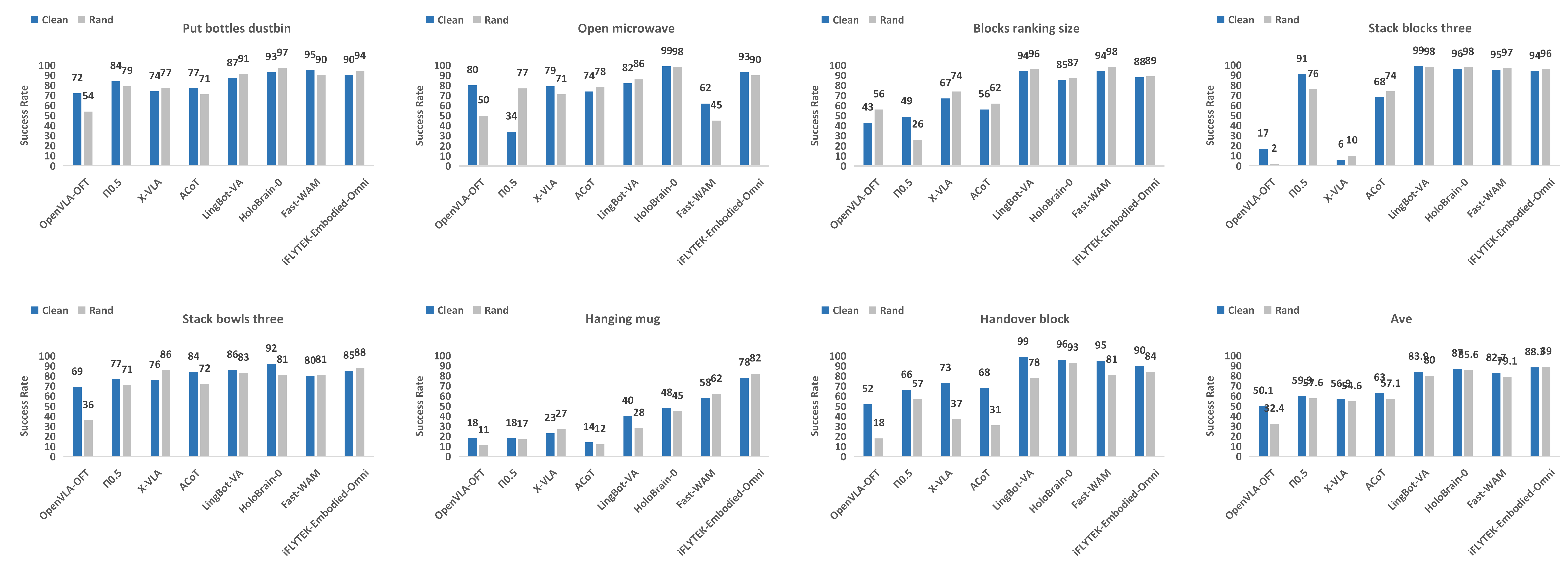}
    \caption{
    Success rates on seven long-horizon tasks selected from RoboTwin 2.0 under the Clean and Rand settings.
    The final panel reports the average performance across all tasks.
    }
    \label{fig:long_horizon_results}
\end{figure*}

As shown in Fig.~\ref{fig:long_horizon_results}, iFLYTEK-Embodied-Omni achieves the highest average success rate in both settings, reaching 88.3\% on Clean and 89.0\% on Rand. These results exceed the second-best averages of HoloBrain-0, which obtains 87.0\% and 85.6\%, by 1.3 and 3.4 percentage points, respectively. Notably, the average success rate increases by 0.7 percentage points under Rand, indicating that environment randomization does not lead to an overall performance degradation across these long-horizon tasks.

The largest advantage appears on Hang Mug, where our model reaches 78\% on Clean and 82\% on Rand, outperforming the corresponding second-best results from Fast-WAM by 20 percentage points in both settings. On Stack Three Bowls, the model achieves the highest Rand success rate of 88\%, exceeding X-VLA by 2 percentage points. It also maintains strong performance on Open Microwave at 93\%/90\%, Put Bottles in Dustbin at 90\%/94\%, and Handover Block at 90\%/84\% under Clean/Rand, respectively.

Although different baselines remain stronger on several individual tasks, iFLYTEK-Embodied-Omni provides the best performance when averaged across the full long-horizon suite. This balanced behavior is consistent with its brain--cerebellum collaborative architecture, in which task decomposition, future-state prediction, progress tracking, and action-chunk generation are coordinated through the shared Omni context.

\subsection{Ablation Study}
\label{sec:ablation}

We conduct two ablation studies on LIBERO-Plus to examine the effectiveness of the three-branch MoT architecture and the proposed multi-view representation. The first study compares the Three-branch MoT design, which assigns separate modeling capacity to the VLM, VGM, and AGM, with a Two-branch MoT variant that combines video and action modeling. The second study compares our view-aware latent fusion with a direct picture-merging strategy. Table~\ref{tab:ablation} reports the success rates under all seven LIBERO-Plus variations.

\begin{table*}[t]
    \centering
    \caption{
    Ablation results on LIBERO-Plus.
    The upper panel evaluates the MoT architecture, and the lower panel evaluates multi-view input encoding.
    The best result within each panel is shown in \textbf{bold}.
    }
    \label{tab:ablation}
    \scriptsize
    \setlength{\tabcolsep}{4pt}
    \renewcommand{\arraystretch}{1.08}
    \resizebox{\textwidth}{!}{
    \begin{tabular}{lcccccccc}
        \hline
        Method &
        Camera &
        Robot &
        Language &
        Light &
        Background &
        Noise &
        Layout &
        Ave. \\
        \hline
        \multicolumn{9}{c}{\textit{MoT architecture}} \\
        \hline
        Two-branch MoT &
        80.1 & 81.9 & 80.6 & 95.5 & 91.3 & 87.9 & 88.2 & 85.9 \\
        iFLYTEK-Embodied-Omni \\
        (Three-branch MoT) &
        \textbf{84.8} & \textbf{85.5} & \textbf{84.6} & \textbf{98.4} &
        \textbf{94.3} & \textbf{92.4} & \textbf{91.2} & \textbf{89.6} \\
        \hline
        \multicolumn{9}{c}{\textit{Multi-view encoding}} \\
        \hline
        Picture-Merge &
        83.5 & 85.2 & 83.3 & 98.1 & 93.9 & 91.2 & 90.4 & 88.8 \\
        iFLYTEK-Embodied-Omni \\
        (View-aware Latent Fusion) &
        \textbf{84.8} & \textbf{85.5} & \textbf{84.6} & \textbf{98.4} &
        \textbf{94.3} & \textbf{92.4} & \textbf{91.2} & \textbf{89.6} \\
        \hline
    \end{tabular}}
\end{table*}

\paragraph{Three-branch MoT architecture.}
Replacing the Two-branch MoT with separate VLM, VGM, and AGM branches improves the average success rate from 85.9\% to 89.6\%, corresponding to a gain of 3.7 percentage points. The Three-branch MoT improves performance under every evaluated variation, with particularly large gains of 4.7 points for Camera, 4.5 points for Noise, and 4.0 points for Language. These results indicate that assigning dedicated Transformer capacity to semantic reasoning, visual dynamics, and action generation reduces interference among their heterogeneous objectives while retaining cross-modal interaction through the shared Omni attention layers.

\begin{figure*}[t]
    \centering
    \begin{minipage}[t]{0.49\textwidth}
        \centering
        \includegraphics[width=\linewidth]{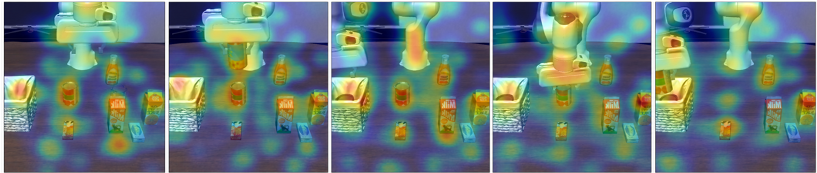}\\[-1mm]
        {\scriptsize (a) Soup-and-sauce task: Two-branch MoT}
    \end{minipage}
    \hfill
    \begin{minipage}[t]{0.49\textwidth}
        \centering
        \includegraphics[width=\linewidth]{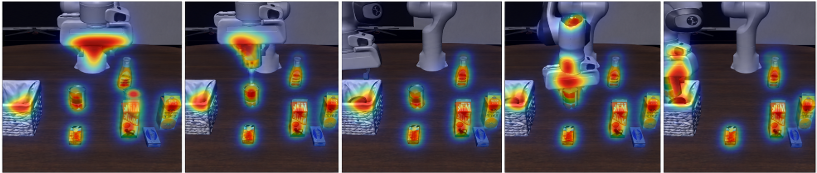}\\[-1mm]
        {\scriptsize (b) Soup-and-sauce task: Three-branch MoT}
    \end{minipage}

    \vspace{1.5mm}

    \begin{minipage}[t]{0.49\textwidth}
        \centering
        \includegraphics[width=\linewidth]{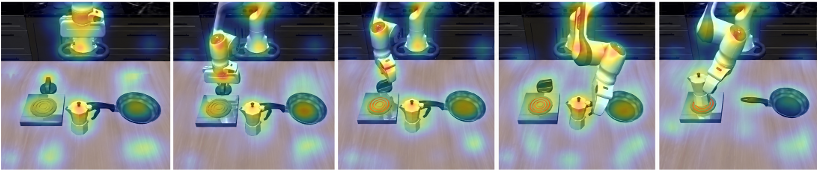}\\[-1mm]
        {\scriptsize (c) Stove-and-moka-pot task: Two-branch MoT}
    \end{minipage}
    \hfill
    \begin{minipage}[t]{0.49\textwidth}
        \centering
        \includegraphics[width=\linewidth]{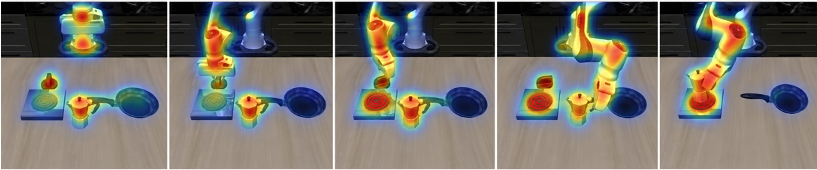}\\[-1mm]
        {\scriptsize (d) Stove-and-moka-pot task: Three-branch MoT}
    \end{minipage}
    \caption{
    Attention-map comparison between Two-branch and Three-branch MoT architectures on two multi-stage manipulation tasks.
    The Three-branch MoT produces more concentrated responses around target objects, the robot end effector, and interaction regions, whereas the Two-branch variant distributes more attention over background areas and task-irrelevant objects.
    }
    \label{fig:mot_attention_ablation}
\end{figure*}

The qualitative comparisons in Fig.~\ref{fig:mot_attention_ablation} provide further evidence for this behavior. In the soup-and-sauce task, the Three-branch MoT more clearly tracks the manipulated container, target items, and robot--object interaction regions across execution stages. In the stove-and-moka-pot task, its responses remain concentrated around the stove control, moka pot, and end effector. In contrast, the Two-branch variant exhibits more diffuse activation over the tabletop and surrounding objects. This more operation-centered attention is consistent with the quantitative improvements observed in Table~\ref{tab:ablation}.

\paragraph{View-aware multi-view encoding.}
We next evaluate how multi-view observations are represented by the VGM. The Picture-Merge baseline concatenates camera images in pixel space before applying a shared VAE encoder and RoPE. Our method instead encodes each camera view independently with the VAE, concatenates the resulting latents, applies RoPE within each view, and adds an independent learnable view embedding to identify the corresponding camera. This view-aware representation improves the average success rate from 88.8\% to 89.6\%. The largest gains are observed for Camera and Language, both increasing by 1.3 percentage points, followed by Noise at 1.2 points and Layout at 0.8 points. These improvements suggest that preserving intra-view spatial structure and camera identity facilitates cross-view correspondence and robust multimodal reasoning.

\section{Limitations and Conclusion}
\label{sec:conclusion}

\textbf{Limitations.}
Despite the effectiveness of the unified architecture, several limitations remain. First, although DiT velocity caching and V2A-style staged denoising enable \textbf{iFLYTEK-Embodied-Omni} to achieve an inference rate of 3 Hz on a single NVIDIA RTX 4090 GPU while maintaining a robot control frequency of 30 Hz, the Three-branch MoT architecture still requires the execution and coordination of the VLM, VGM, and AGM computation paths. Its inference cost therefore leaves room for improvement in applications requiring lower latency, higher-frequency control, or deployment on resource-constrained robot platforms. Model compression, dynamic expert routing, and more aggressive reuse of intermediate computations may further improve efficiency. Second, the action-chunk length is currently specified by a fixed hyperparameter and cannot adapt to the duration or complexity of subgoals. A fixed chunk may generate redundant controls for short operations, while long or difficult subtasks may require unnecessarily frequent replanning. Future work should investigate subgoal-conditioned adaptive action chunking that dynamically determines the prediction horizon from task context and environment feedback.

\textbf{Conclusion.}
We presented \textbf{iFLYTEK-Embodied-Omni}, a unified multimodal foundation model that jointly models vision, language, video, and action within a single Omni framework. The model establishes a brain--cerebellum organization in which the VLM and VGM perform semantic reasoning, task planning, progress tracking, and visual world modeling, while the AGM converts planned subgoals into executable action chunks. A multi-source training mixture connects general multimodal knowledge, embodied perception and reasoning, action-free interaction videos, and action-labeled robot trajectories. We further introduced a four-stage progressive-to-joint training strategy that first develops the specialized capabilities of the VLM, VGM, and AGM and subsequently aligns them through end-to-end optimization. Experiments on LIBERO-Plus, RoboTwin 2.0, and long-horizon manipulation tasks demonstrate strong generalization and robustness, while the ablation studies validate the effectiveness of the Three-branch MoT architecture and view-aware multi-view encoding.

Looking forward, we plan to incorporate speech as an additional modality within the Omni framework. Jointly aligning speech, vision, language, video, and action would allow robots to understand spoken instructions, request clarification when task intent is ambiguous, and provide verbal feedback about execution progress. Such an extension could enable more natural, low-latency, and seamless communication between humans and embodied agents.

\clearpage

% \bibliographystyle{plainnat}
% \bibliography{main}

\bibliographystyle{plainnat}
\bibliography{references}

\end{document}